\documentclass[lettersize,journal]{IEEEtran}
\usepackage{amsmath,amsfonts}
\usepackage{algorithmic}
\usepackage{algorithm}
\usepackage{array}
\usepackage{textcomp}
\usepackage{stfloats}
\usepackage{url}
\usepackage{verbatim}
\usepackage{graphicx}
\usepackage{cite}
\usepackage{pgfplots}

\usepackage{array}
\usepackage{lineno}
\usepackage{graphicx}
\usepackage{epstopdf}
\usepackage{xcolor}
\usepackage{amssymb,amsfonts,amsmath,amscd,mathrsfs,bm}
\usepackage{fancyhdr}
\usepackage{marvosym}
\usepackage{amsthm}
\usepackage{amsmath}
\usepackage{bm}
\usepackage{booktabs}
\usepackage{multirow}
\usepackage{makecell}
\usepackage{hhline}
\usepackage{stmaryrd}
\usepackage{color, xcolor}
\usepackage{dcolumn}
\usepackage{algorithm}
\usepackage{algorithmic}
\usepackage{lipsum}
\usepackage[mathscr]{eucal}

\usepackage{mathtools}
\usepackage{cite}
\usepackage{balance}
\usepackage{float}
\usepackage{hyperref}
\usepackage{gensymb}
\usepackage{verbatim}
\usepackage{setspace}
\newcommand{\tabincell}[2]{\begin{tabular}{@{}#1@{}}#2\end{tabular}}
\begin{document}

\title{A Domain-adaptive Physics-informed Neural Network for Inverse Problems of Maxwell's Equations in Heterogeneous Media}

\author{Shiyuan Piao, Hong Gu, Aina Wang, Pan Qin
\thanks{The authors are with the Key Laboratory of Intelligent Control and Optimization for Industrial Equipment of Ministry of Education and the School of Control Science and Engineering, Dalian University of Technology, Dalian 116024, China e-mail: sewon@mail.dlut.edu.cn, guhong@dlut.edu.cn, WangAn@mail.dlut.edu.cn, qp112cn@dlut.edu.cn (\emph{Corresponding author: Pan Qin})}
\thanks{}}


\maketitle

\begin{abstract}
Maxwell's equations are a collection of coupled partial differential equations (PDEs) that, together with the Lorentz force law, constitute the basis of classical electromagnetism and electric circuits.
Effectively solving Maxwell's equations is crucial in various fields, like electromagnetic scattering and antenna design optimization.
Physics-informed neural networks (PINNs) have shown powerful ability in solving PDEs. 
However, PINNs still struggle to solve  Maxwell's equations in heterogeneous media. 
To this end,  we propose a domain-adaptive PINN (da-PINN) to solve inverse problems of Maxwell's equations in heterogeneous media.
First, we propose a location parameter of media interface to decompose the whole domain into several sub-domains.
Furthermore, the electromagnetic interface conditions are incorporated into a loss function to improve the prediction performance near the interface.
Then, we propose a domain-adaptive training strategy for da-PINN.
Finally, the effectiveness of da-PINN is verified with two case studies.\end{abstract}


\section{Introduction}
\IEEEPARstart{M}{axwell's} equations are used to describe the fundamental laws of electromagnetic fields \cite{EBRAHIMIJAHAN2022397}, which are widely used in geological exploration \cite{4540855}, medical imaging \cite{9325009},  and many other fields \cite{6710928,9195159}.
The Maxwell's equations are classically solved by numerical methods \cite{9665310}, such as finite-difference time-domain method \cite{ZHAO200460}, finite element method \cite{doi:10.1137/0729045}, Born iterative method \cite{1993Study}, and variational Born iteration method \cite{2000Variational}.
However,  one of the challenges these numerical methods face is the high computational cost\cite{10068421}.

The deep learning methods with remarkable approximation capability have attracted increasing attention in solving partial differential equations (PDEs) \cite{9428581, molinaro2023neural}.
The physics-informed neural networks (PINNs) proposed in \cite{raissi2017physics} are prevalent mesh-free methods for solving PDEs.
In the framework of PINNs, the prior knowledge of physics and sparse measurements are incorporated into the loss function to train PINNs. 
Successful achievements in estimating the parameters of PDEs with homogeneous media, such as supersonic flows \cite{JAGTAP2022111402},  beam systems \cite{kapoor2023physicsinformed}, and nano-optics \cite{Chen_2020}, have been reported \cite{ye2021deep,kapoor2023load}.  
Meanwhile, the Maxwell's equations in heterogeneous media have wide applications, such as nondestructive testing of multilayered coating\cite{4352036,9632211}.
However, it remains a challenge for PINNs to solve inverse problems in heterogeneous media composed of several homogeneous media \cite{faroughi2023physicsguided}. 
In the heterogeneous media case, the parameters of PDEs are not fixed in the whole domain and will jump at media interface. 
Such parameter jumping can potentially lead to PINNs could not work well.
In\cite{chen2021physicsinformed,9709153}, PINNs were used to solve inverse problems of the Maxwell's equations in heterogeneous media.
However, these methods obtain inaccurate estimations near the interface due to the interface conditions not being considered, as indicated in our experiments.

To solve the aforementioned problems, this letter proposes a domain-adaptive PINN (da-PINN) for the Maxwell's equations in heterogeneous media.
The goal of da-PINN is to estimate parameters of the Maxwell's equations and predict the electric and magnetic fields.
We first propose a location parameter of media interface to decompose the whole domain into several sub-domains under the assumption of each sub-domain with fixed parameters of the Maxwell's equations.
Then, sub-networks are constructed to approximate electric and magnetic fields in each sub-domain.
The electromagnetic interface conditions are incorporated into a loss function.
In addition, a domain-adaptive training strategy is proposed to optimize da-PINN.
Finally, the performance of da-PINN is verified with one- and two- dimensional Maxwell's equations in heterogeneous media.

This letter is organized as follows. In Section \ref{se2}, we introduce PINNs for a derivative of vector-valued functions and da-PINN. In Section \ref{se3},  da-PINN is verified with one- and two- dimensional cases. Section \ref{se4} concludes this letter.
\section{Methodology}\label{se2}
\subsection{PINNs for Derivatives of Vector-valued Functions}\label{PINNs}
In this section, we briefly introduce PINNs for inverse problems in the derivative of vector-valued functions, the general form is as follows: 
\begin{equation}\label{eq:PDE1}
        \boldsymbol{u}_{t}(t,\boldsymbol{x})+\boldsymbol{\mathcal{N}}\left[ \boldsymbol{u}(t,\boldsymbol{x});\boldsymbol{\lambda}  \right]=0,~\boldsymbol{x}\in \Omega \subseteq \mathbb{R}^{D} ,~t\in \left[ 0,T \right].
\end{equation}
Here, $\boldsymbol{u}:\mathbb{R}\times\mathbb{R}^{p}\to \mathbb{R}^{q}$ denotes the solution with  spatiotemporal dependence, $\boldsymbol{u}_{t}$ is the partial derivative of  $\boldsymbol{u}$ with respect to $t$, $\boldsymbol{\mathcal{N}}$ is a vector differential operator parameterized by an unknown real-valued vector $\boldsymbol{\lambda}\in \mathbb{R}^{m}$, $(t,\boldsymbol{x})$ is a pair-wised coordinate of the time $t$ and space $\boldsymbol{x}$.
A residual function $\boldsymbol{f}(t, \boldsymbol{x})$ is defined as follows:
\begin{equation}\label{eq:f1}
    \boldsymbol{f} =\boldsymbol{u}_{t}+\boldsymbol{\mathcal{N}}\left[ \boldsymbol{u};\boldsymbol{\lambda}  \right].
\end{equation}
In PINNs, $\widehat{\boldsymbol{u}}\left( t,\boldsymbol{x};\boldsymbol{\theta } \right)$ is a fully-connected neural network (FCNN) to approximate $\boldsymbol{u}$ satisfying \eqref{eq:PDE1}, in which $\boldsymbol{\theta}$ is a set of weights.
PINNs can be trained using the following loss function:
\begin{equation}\label{eq:loss1}
        Loss(\boldsymbol{\theta},\boldsymbol{\lambda}) = Loss_{D}(\boldsymbol{\theta}) + Loss_{P}(\boldsymbol{\theta},\boldsymbol{\lambda}),
\end{equation}
where
\begin{equation}\notag
        Loss_{D}(\boldsymbol{\theta}) = \dfrac{1}{N} \sum_{(t,\boldsymbol{x},\boldsymbol{u})\in D_D}\left\| \widehat{\boldsymbol{u}}\left (t,\boldsymbol{x};\boldsymbol{\theta }  \right ) -\boldsymbol{u} \right\|_2^{2}
\end{equation}        
is a data-driven loss;
\begin{equation}\notag
        Loss_{P}(\boldsymbol{\theta},\boldsymbol{\lambda}) = \dfrac{1}{N} \sum_{(t,\boldsymbol{x})\in D_C} \left\|\widehat{\boldsymbol{f}}\left(t,\boldsymbol{x};\boldsymbol{\theta },\boldsymbol{\lambda}  \right)\right\|_2 ^{2}  
\end{equation}
is a physics-informed loss;
$D_D=\{(t_{j},\boldsymbol{x}_{j},\boldsymbol{u}_{j})|j=1,2,\ldots,N\}$ denotes the training dataset; 
$D_C=\{(t_{j},\boldsymbol{x}_{j})|j=1,2,\ldots,N\}$ denotes the collocation point set.
Unknown $\boldsymbol{\lambda}$ is treated as weights of neural networks and will be optimized by the gradient descent method. 
The automatic differentiation (AD) \cite{baydin2018automatic} is used to numerically calculate $\boldsymbol{u}_{t}$ and $\mathcal{N}\left[ \boldsymbol{u};\boldsymbol{\lambda}  \right]$.

\subsection{da-PINN for Maxwell's Equations in Heterogeneous Media}\label{da-PINN}
The passive and lossless Maxwell's equations under study are of the following generalized form:
\begin{equation}\label{eq:maxwellEquation2}
    \begin{array}{ll}
    \nabla  \times \boldsymbol{E}(t,\boldsymbol{x})=-\mu\left ( \boldsymbol{x} \right ) \dfrac{\partial \boldsymbol{H}(t,\boldsymbol{x}) }{\partial t},\\
    \nabla  \times \boldsymbol{H}(t,\boldsymbol{x})=\varepsilon\left ( \boldsymbol{x} \right ) \dfrac{\partial \boldsymbol{E}(t,\boldsymbol{x})}{\partial t},
    \end{array}
\end{equation}
with
\begin{equation}\notag
    \begin{array}{ll}
    \mu\left ( \boldsymbol{x} \right ) =\begin{cases}
        \mu_{1},  & \text{ if } \boldsymbol{x}\in \Omega _{1} \\
        \mu_{2},   & \text{ if } \boldsymbol{x}\in \Omega _{2}
    \end{cases},
    \hspace{1.0em}
    \varepsilon \left ( \boldsymbol{x} \right ) =\begin{cases}
        \varepsilon_{1},  & \text{ if } \boldsymbol{x}\in \Omega _{1} \\
        \varepsilon_{2},   & \text{ if } \boldsymbol{x}\in \Omega _{2}
        \end{cases}
    \end{array}.
\end{equation}
Here, $\boldsymbol{E}$ is the electric field vector, $\boldsymbol{H}$ is the magnetic field vector,
$\mu$ is the magnetic permeability, $\varepsilon$ is the electric permittivity, $\Omega$ is a spatial close set with heterogeneous media, which is composed of two homogeneous media.
$\nabla$ denotes the del operator, and $\times$ denotes the cross product.
$\Omega$ is divided into $\Omega _{1} \cup \Omega _{2}$. The interface $\psi =\partial \Omega _{1} \cap \partial\Omega _{2} $ with $\partial$ being the boundary.

According to \cite{Cheng2007Field}, the electromagnetic interface conditions under study are as follows: 
\begin{equation}\label{eq:interface}\notag
    \begin{cases}
        E _{1t}  =E _{2t},
        \hspace{12mm}
        H _{1t}  =H _{2t}\\
        \varepsilon_{1}E_{1n}  =\varepsilon_{2}E_{2n},
        \hspace{5mm}
        \mu_{1}H_{1n}  =\mu_{2}H_{2n}
    \end{cases}
        ,
\end{equation}
where $E_{it}$ and $E_{in}$ are the tangential and normal components of electric fields in $\Omega _{i}$, respectively.
$ H_{it}$ and $H_{in}$ are the tangential and normal components of  magnetic fields in $\Omega _{i}$, respectively.
In this letter, $i=1,2$.
Residual functions for the Maxwell's equations and interface conditions are defined as follows:
\begin{align} 
    &\boldsymbol{f} = \nabla  \times \boldsymbol{E}+\mu \dfrac{\partial \boldsymbol{H}}{\partial t}, \label{eq:Mf1}\\
    &\boldsymbol{h} =\nabla  \times \boldsymbol{H} - \varepsilon \dfrac{\partial \boldsymbol{E} }{\partial t},\label{eq:Mh1} \\
    &s = \left( E _{1t} -E_{2t} \right)^{2} +\left( H_{1t}-H_{2t} \right)^{2}\notag\\
    &+ \left( \varepsilon_{1}E _{1n}-\varepsilon_{2}E_{2n} \right)^{2}+\left( \mu_{1}H_{1n}-\mu_{2}H_{2n} \right)^{2}.\label{eq:f4}
\end{align}

Assume that  $\psi$ in  $\Omega$ is perpendicular to the $x$-axis, but the exact location of $\psi$, i.e., $x=d$ is unknown.
To this end, da-PINN is proposed to estimate the parameter vector $\boldsymbol{\lambda} =\left[\mu_{1}, \varepsilon_{1}, \mu _{2}, \varepsilon_{2}, d\right]^\top$ and  predict $\boldsymbol{E}$ and $\boldsymbol{H}$;
da-PINN can be trained using the following loss function:
\begin{equation}\label{eq:lossme}
    \begin{aligned}
    Loss(\boldsymbol{\theta_1},\boldsymbol{\theta_2},\boldsymbol{\lambda})=&Loss_{D}(\boldsymbol{\theta_1},\boldsymbol{\theta_2})+ Loss_{P}(\boldsymbol{\theta_1},\boldsymbol{\theta_2},\boldsymbol{\lambda})\\ &+ Loss_{I}(\boldsymbol{\theta_1},\boldsymbol{\theta_2},\boldsymbol{\lambda}),
    \end{aligned}
\end{equation}
where
\begin{equation}\notag
        \begin{aligned}
        Loss_{D}(\boldsymbol{\theta_1},\boldsymbol{\theta_2}) = &\dfrac{1}{N_{D_1}} \sum_{(t,\boldsymbol{x},u)\in D_{D_1}}\left\| \widehat{\boldsymbol{u}}_{1}\left(t,\boldsymbol{x};\boldsymbol{\theta }_{1}  \right) -\boldsymbol{u}\right\|_2^{2}\\[1mm]
      &+\dfrac{1}{N_{D_2}} \sum_{(t,\boldsymbol{x},u)\in D_{D_2}}\left\| \widehat{\boldsymbol{u}}_{2}\left(t,\boldsymbol{x};\boldsymbol{\theta }_{2}  \right) -\boldsymbol{u}  \right\|_2^{2}\\[3mm]
        \end{aligned}
\end{equation}
is a data-driven loss;

\begin{equation}\notag
        \begin{aligned}
        &Loss_{P}(\boldsymbol{\theta_1},\boldsymbol{\theta_2},\boldsymbol{\lambda}) = \dfrac{1}{N_{P_1}}\sum_{(t,\boldsymbol{x})\in C_{P_1}}\hspace{-3mm}\left\{ \left\|\widehat{\boldsymbol{f}}_{1}\left(t,\boldsymbol{x};\boldsymbol{\theta }_{1},\boldsymbol{\lambda}  \right)\right\|_2^{2}\right.\\ &+ \left.\|\widehat{\boldsymbol{h}}_{1}(t,\boldsymbol{x};\boldsymbol{\theta }_{1},\boldsymbol{\lambda}  )\|_2^{2}\right\}+ \dfrac{1}{N_{P_2}} \sum_{(t,\boldsymbol{x})\in C_{P_2}}\hspace{-3mm}\left\{\left\|\widehat{\boldsymbol{f}}_{2}\left(t,\boldsymbol{x};\boldsymbol{\theta }_{2} ,\boldsymbol{\lambda} \right)\right\|_2^{2}\right. \\ &+\left.\left\|\widehat{\boldsymbol{h}}_{2}\left(t,\boldsymbol{x};\boldsymbol{\theta }_{2},\boldsymbol{\lambda}  \right)\right\|_2^{2}\right\}\\[3mm]
        \end{aligned}
\end{equation}
is a physics-informed loss;
\begin{equation}\notag
        Loss_{I}(\boldsymbol{\theta_1},\boldsymbol{\theta_2},\boldsymbol{\lambda}) = \dfrac{1}{N_I} \sum_{(t,\boldsymbol{x})\in C_{I}}\widehat{s}\left(t,\boldsymbol{x};\boldsymbol{\theta }_{1},\boldsymbol{\theta }_{2},\boldsymbol{\lambda}  \right) ^{2}
\end{equation}
is an interface-condition loss; 
$\widehat{\boldsymbol{u}}_{i}\left( t,\boldsymbol{x};\boldsymbol{\theta}_{i} \right)=\left[\widehat{\boldsymbol{E}}_{i}(t,\boldsymbol{x};\boldsymbol{\theta}_i)^\top,\widehat{\boldsymbol{H}}_{i}(t,\boldsymbol{x};\boldsymbol{\theta}_i)^\top\right]^\top$ is the function of the constructed $Net_i$ with $\boldsymbol{\theta }_{i} $ being a set of weights.
$D_{D}=D_{D_1}\cup D_{D_2}$, $D_{D_i}=\{(t_{j},\boldsymbol{x}_{j},u_{j})|j=1,2,\ldots,N_{D_i}\}$ and $C_{P_i}=\left\{(t_{j},\boldsymbol{x}_{j})\left|j=1,2,\ldots,N_{P_i}\right.\right\} $ are the training dataset and collocation point sets, respectively. They are randomly sampled from the whole domain.
$C_{I}=\{(t_{j},\boldsymbol{x}_{j})|j=1,2,\ldots,N_{I}\} $ is the collocation point set randomly sampled from the interface; 
$\boldsymbol{\hat f}_{i},\boldsymbol{\hat h}_{i}$, and ${\hat s}$ are obtained according to \eqref{eq:Mf1}-\eqref{eq:f4}.
\par
A domain-adaptive training strategy shown in Algorithm  \ref{alg:algorithm-label} is proposed to achieve  $\boldsymbol{\widehat{\theta}_i}$ and $\widehat{\boldsymbol{\lambda}}$ simultaneously.
Let $k$ denote iterative step.
In Algorithm \ref{alg:algorithm-label}, 
we take $\nu^{(k)}  d^{(k)}$ as the $x$-axis coordinate of $C_{P_1}^{(k)}$, $\nu^{(k)}  \left(B-d^{(k)}\right)+d^{(k)}$ as the $x$-axis coordinate of $C_{P_2}^{(k)}$, and $d^{(k)}$ as the $x$-axis coordinate of $C_{I}^{(k)}$.
Here, $\nu^{(k)}\overset{{i.i.d.}} {\sim} \mathrm{U}\left(0,1\right)$ and $x \in \left[0,B\right]$  with $B$ being an upper bound of $x$.
Note that updating $d^{(k)}$ can constantly change $D_{D_i}^{(k)}$,  $C_{P_i}^{(k)}$, and $C_{I}^{(k)}$ in  \eqref{eq:lossme}
.
\begin{algorithm}
    \caption{The domain-adaptive training strategy of optimizing for da-PINN}
    \label{alg:algorithm-label}
    \begin{algorithmic}
    \STATE -\textbf{Inputs}: $(t,\boldsymbol{x})$. 
    \STATE -\textbf{Initialize}: Randomly initialize $\boldsymbol{\theta} ^{(0)}_{i}$ for $Net_i$ and parameters  $\boldsymbol{\lambda}^{(0)}$; $\left(k=0\right)$.
    \WHILE {the stop criterion is not satisfied }
    \STATE 1) Based on the parameter of interface location $d^{(k)}$, divide training datasets $D_{D}$ into $D_{D_i}^{(k)}$ and randomly sample collocation point sets $C_{Pi}^{(k)}$ and $C_{I}^{(k)}$.
    \STATE 2) Update $\boldsymbol{\theta}^{(k)}_{i}$ and  $\boldsymbol{\lambda}^{(k)}$ according to $Loss^{(k)}$. 
    \begin{equation}\notag
        \begin{cases}
            \boldsymbol{\theta}_{i}^{\left (  k+1\right ) } = \boldsymbol{\theta}_{i}^{\left (  k\right ) } -\eta \nabla _{\boldsymbol{\theta}_{i}}Loss^{(k)}\\[2mm]
            \boldsymbol{\lambda}^{(k+1)} = \boldsymbol{\lambda}^{\left(  k\right)} -\eta\nabla _{\boldsymbol{\lambda}}Loss^{(k)}
            \end{cases}.
    \end{equation}
    \STATE 3) $k=k+1$.
    \ENDWHILE
    \STATE -\textbf{Return}: The output $\widehat{\boldsymbol{u}}_{i}\left( t,\boldsymbol{x};\boldsymbol{\hat\theta}_{i} \right) $ and $\widehat{\boldsymbol{\lambda}}$.
    \end{algorithmic}
\end{algorithm}
\section{Experiments}\label{se3}
In this section, da-PINN is verified with the one- and two-dimensional Maxwell's equations in heterogeneous media and unknown interface location.
All experiments are implemented by Pytorch. Adam is used as the optimizer. The following $l_{2}$ relative error is used to evaluate the performance of da-PINN:
\begin{equation}\notag
    l_{2}\mbox{ relative error} = \dfrac{\left\| \boldsymbol{u}-\widehat{\boldsymbol{u}} \right\|_{2}}{\left\| \boldsymbol{u} \right\|_{2}},\\
\end{equation} 
where $\boldsymbol{u}$ denotes the ground truth and $\widehat{\boldsymbol{u}}$ denotes the corresponding prediction or estimation.
\subsection{One-dimensional Maxwell's Equations}
\begin{figure}[htb]
    \begin{center}
    \includegraphics[width=\linewidth]{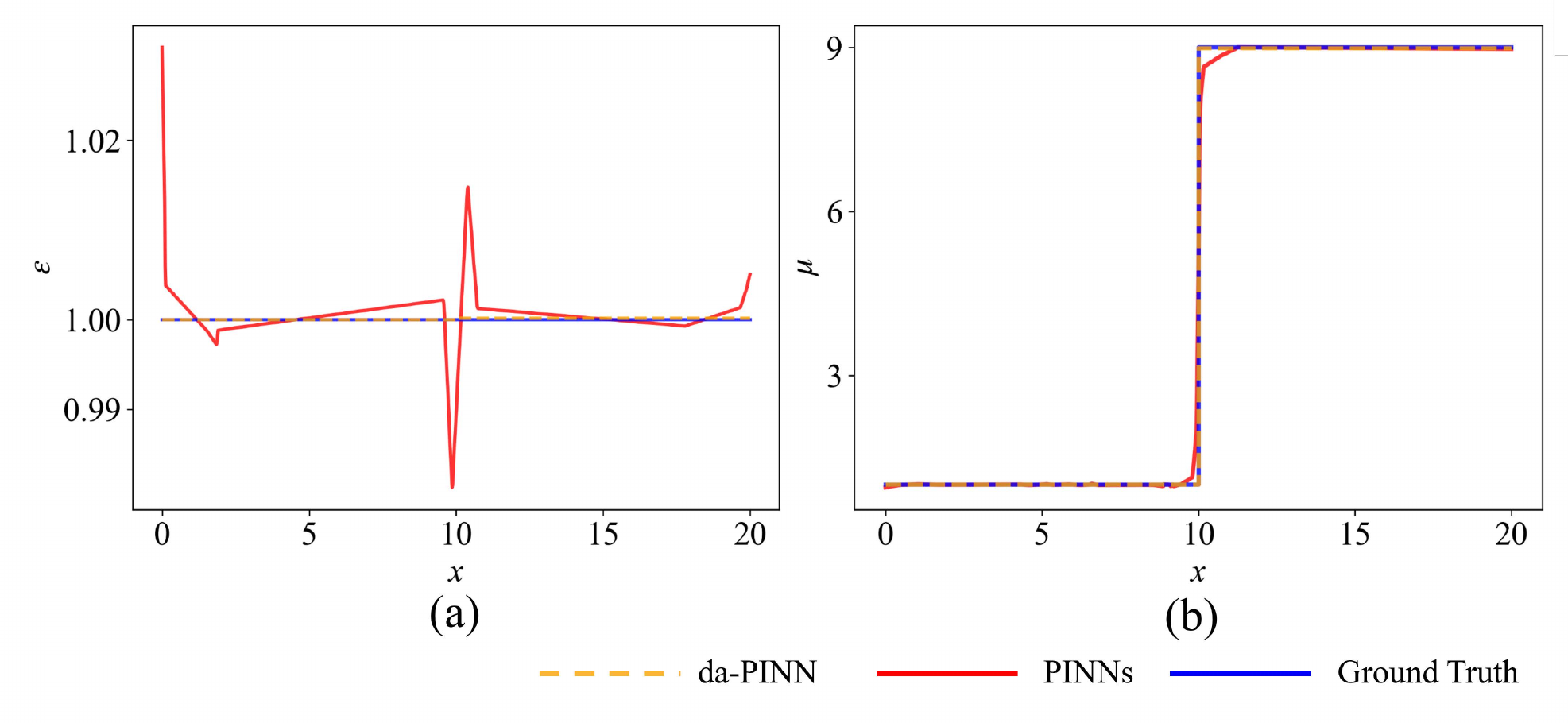}\vspace{-3mm}
        \caption{Estimations obtained by da-PINN and PINNs. (a) Estimations for $\varepsilon$. (b) Estimations for $\mu$.}
    \vspace{-7mm}
    \label{fig:1Dinverse}
    \end{center}
\end{figure}

\begin{figure}[htb]
    \begin{center}
    \includegraphics[width=7cm]{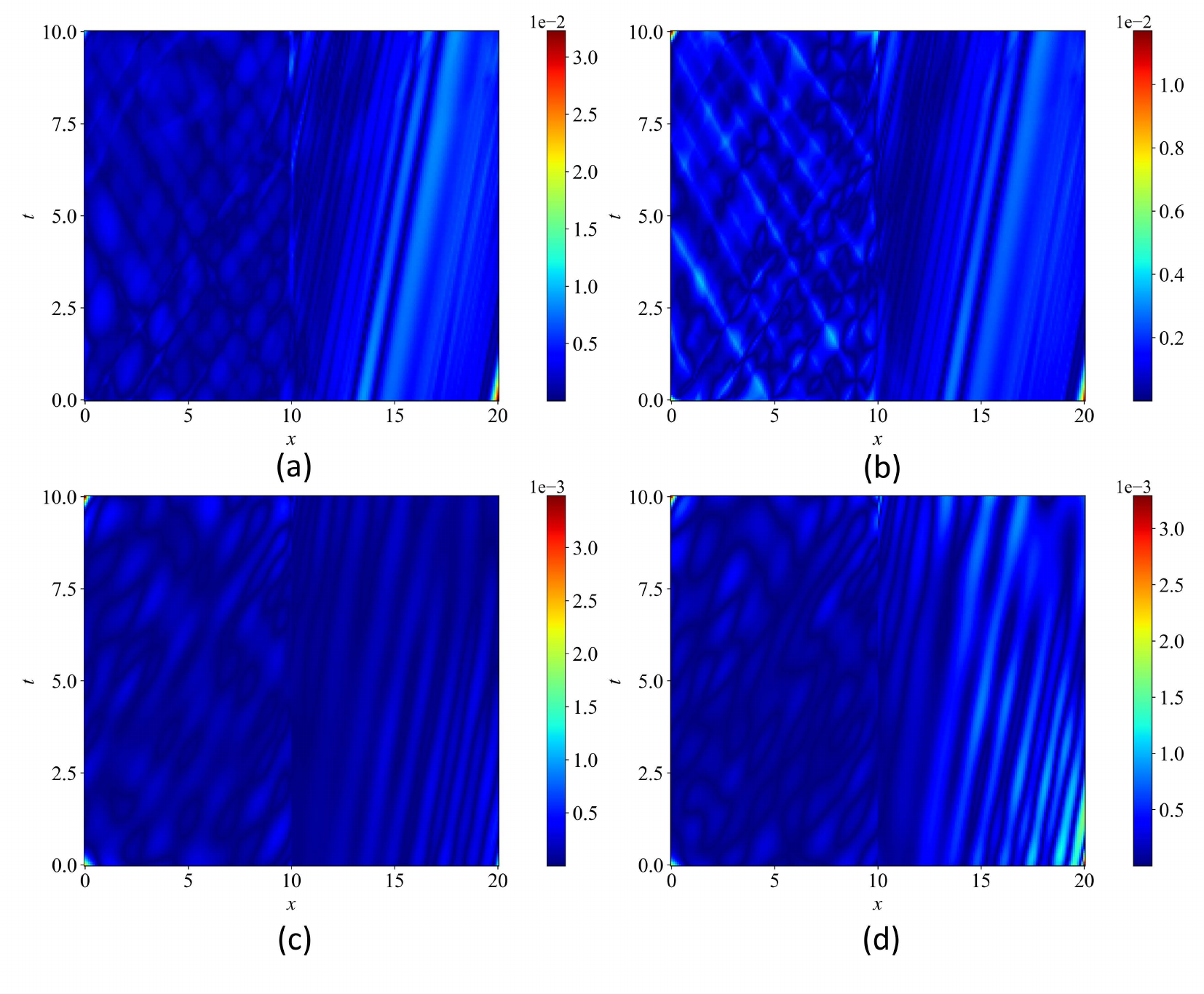}  
    \vspace{-3mm}
    \caption{Absolute errors for predictions to one-dimensional Maxwell's equations. (a) $E_{Y}$ obtained  by da-PINN.  (b) $H_{Z}$ obtained by da-PINN. (c) $E_{Y}$ obtained by PINNs. (d) $H_{Z}$ obtained by PINNs.}
    \vspace{-1mm}
    \label{fig:1D}
    \end{center}
\end{figure}

The one-dimensional Maxwell's equations in heterogeneous media under study are as follows:
\begin{equation}\label{eq:1dmaxwell}
    \hspace{-4mm}\dfrac{\partial E_{Y}}{\partial x} \hspace{-1mm}=\hspace{-1mm} -\mu\dfrac{\partial H_{Z}}{\partial t},
    \dfrac{\partial H_{Z}}{\partial x} \hspace{-1mm}=\hspace{-1mm} -\varepsilon \dfrac{\partial E_{Y}}{\partial t}, x\in \left[ 0,20 \right], t\in \left[ 0,10 \right].
\end{equation}
$\Omega _{1}=\left[ 0,10 \right]$ and $\Omega _{2}=\left[ 10,20 \right]$ are obtained with  $x=10$  as $d$.
$\varepsilon =1$  in the whole $\Omega$.
To present heterogeneous media,  $\mu$ is assumed to be a piecewise constant. 
Here, $\mu_1=1$ in $~\Omega _{1}$ and $\mu_2=9$ in $~\Omega _{2}$.
The analytical solutions to \eqref{eq:1dmaxwell} are as follows:
\begin{eqnarray*}
E_Y = \left\{\begin{array}{ll}
\begin{array}{l}
\!\!\mathrm{cos}\left( 0.1t-0.1x+1 \right)\\
\quad+0.5\mathrm{cos}\left( 0.1t-0.1x-1 \right), 
\end{array} &\text{ if } x \in \Omega _{1} \\[5mm]
1.5\mathrm{cos}\left( 0.1t-0.3x+3 \right), &\text{ if } x \in \Omega _{2}
\end{array}\right.
,
\end{eqnarray*}
\vspace{-3mm}
\begin{eqnarray*}
    H_Z = \left\{\begin{array}{ll}
    \begin{array}{l}
    \!\!\mathrm{cos}\left( 0.1t-0.1x+1 \right)\\
    \quad-0.5\mathrm{cos}\left( 0.1t-0.1x-1 \right), 
    \end{array} &\text{ if } x \in \Omega _{1} \\[5mm]
    0.5\mathrm{cos}\left( 0.1t-0.3x+3 \right), &\text{ if } x \in \Omega _{2}
    \end{array}\right.
,
\end{eqnarray*}
The abovementioned setups are used to generate training and testing datasets. Considering the practical operating environment, $\boldsymbol{\lambda}$ is assumed to be unknown.
In this example, we set $N_{D}=2000$, $N_{P1}=N_{P2}=4000$, and $N_{I}=2000$.
There are $5$ hidden layers and $30$ neurons per layer in $Net_{i}$. 
ReLU is used as the activation function.
Randomly initialize parameters  as $\mu_{1}^{(0)}=1$, $\varepsilon_{1}^{(0)} =1$, $\mu_{2}^{(0)}=13$, $\varepsilon_{2}^{(0)}=0$, and $d^{(0)}=15$.
Table \ref{table:1Dinversmax} shows the estimation performance.
\begin{table}[htb]
    \tabcolsep=1.5cm
	\renewcommand\arraystretch{1}
    \begin{center}
    \vspace{-3mm}
    \caption{estimation performance of da-PINN for one-dimensional Maxwell's equations.}
    \setlength{\tabcolsep}{4mm}{
    \begin{tabular}{c c c c}
    \hline
    \tabincell{c}{Parameters} &
    \tabincell{c}{Estimations} & 
    \tabincell{c}{$l_2$ relative error (\%)} \\\hline
    {\tabincell{c}{${\rm \mu_{1}}$}} & 0.9997 & 0.0309 \\ 
    {\tabincell{c}{${\rm \varepsilon_{1}}$}}  & 0.9998 & 0.0165 \\                              
    {\tabincell{c}{${\rm \mu_{2}}$}}  & 8.9918 & 0.0907 \\ 
    {\tabincell{c}{${\rm \varepsilon_{2}}$}}  & 0.9999 & 0.0132 \\
    {\tabincell{c}{${d}$}}  & 9.9998 & 0.0017  \\\hline
    \end{tabular}}
    \label{table:1Dinversmax}
    \vspace{-2mm}
    \end{center}
\end{table}
The estimations of parameters $\varepsilon$ and $\mu$ obtained by da-PINN are shown in Fig. \ref{fig:1Dinverse}. 
The absolute errors $\left|u-\widehat{u}\right|$ of $E_{Y}$ and $H_{Z}$ obtained by da-PINN and PINNs are shown in Fig. \ref{fig:1D}.
The results indicate that da-PINN achieves better performance, especially near the interface.
The $l_{2}\mbox{ relative errors}$ are listed in Table \ref{table:1Dpred}, in which the best values are in bold font.
\vspace{-1mm}
\begin{table}[htb]
    \tabcolsep=0.5cm
	\renewcommand\arraystretch{1}
    \begin{center}
    \caption{Prediction performance of da-PINN and PINNs for one-dimensional Maxwell's equations.}
    \setlength{\tabcolsep}{3mm}{
    \begin{tabular}{c c c c}
    \hline
    \tabincell{c}{Methods} &
    \tabincell{c}{Predictions} &
    \tabincell{c}{$l_{2}$ relative error}\\\hline
    \multirow{2}*{da-PINN}&  {\tabincell{c}{${\widehat{E}_{Y}}$}} &  $\textbf{6.485}\boldsymbol{e}$-$\textbf{4}$ \\
    ~&  {\tabincell{c}{${\widehat{H}_{Z}}$}} &  $\textbf{1.120}\boldsymbol{e}$-$\textbf{3}$   \\
    \multirow{2}*{PINNs}&  {\tabincell{c}{${\widehat{E}_{Y}}$}} &  $2.664e$-$3$  \\
    ~&  {\tabincell{c}{${\widehat{H}_{Z}}$}} &  $3.627e$-$3$  \\\hline
    \end{tabular}}
    \label{table:1Dpred}
    \end{center}
\end{table}
\subsection{Two-dimensional Maxwell's Equations}
\begin{figure}[htb]
    \begin{center}
    \includegraphics[width=\linewidth]{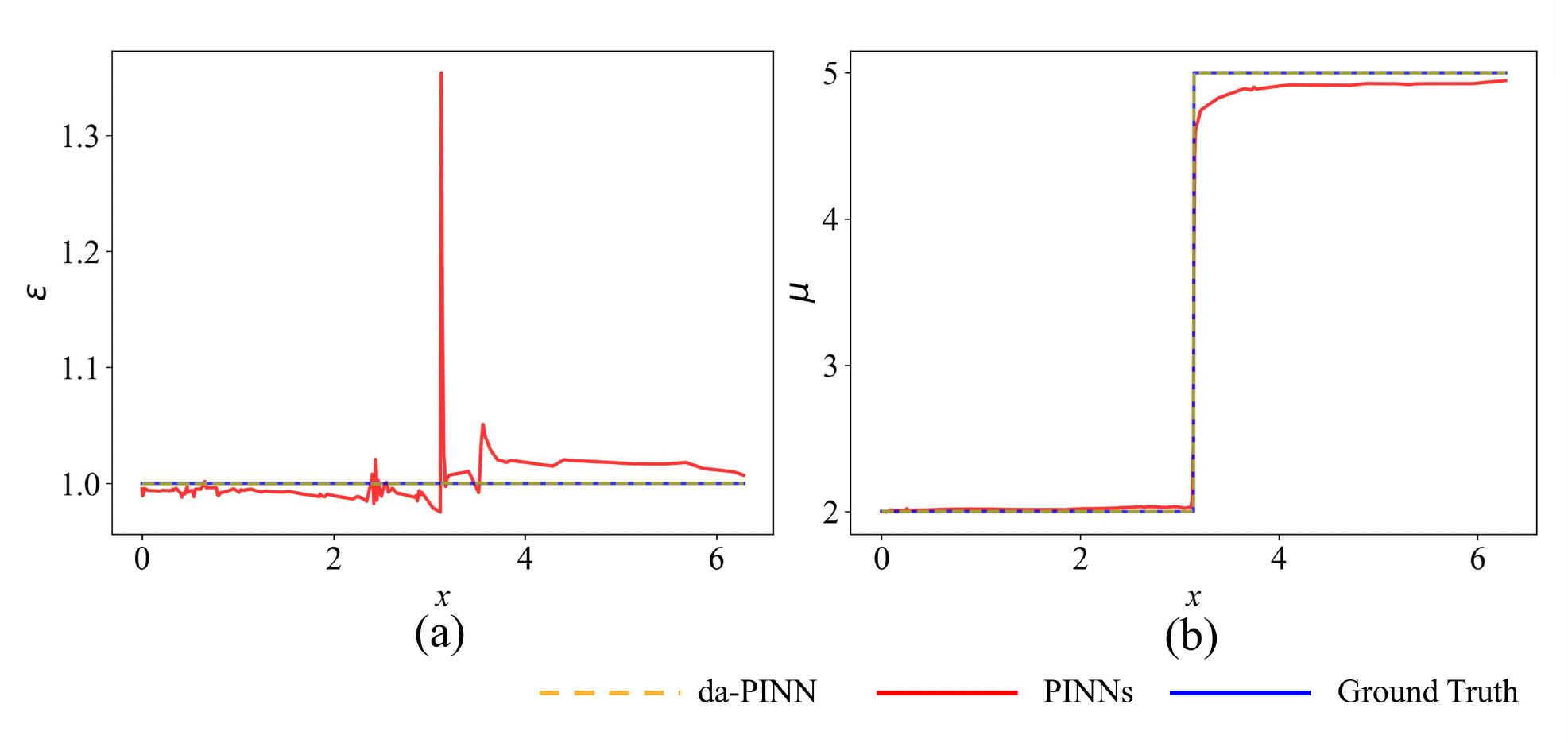}    

    \caption{Estimations obtained by da-PINN and PINNs. (a) Estimations for $\varepsilon$. (b) Estimations for $\mu$.}
    \label{fig:2dinverse}
    \end{center}
\end{figure}

\begin{figure}[htb]
    \begin{center}
    \includegraphics[width=8cm]{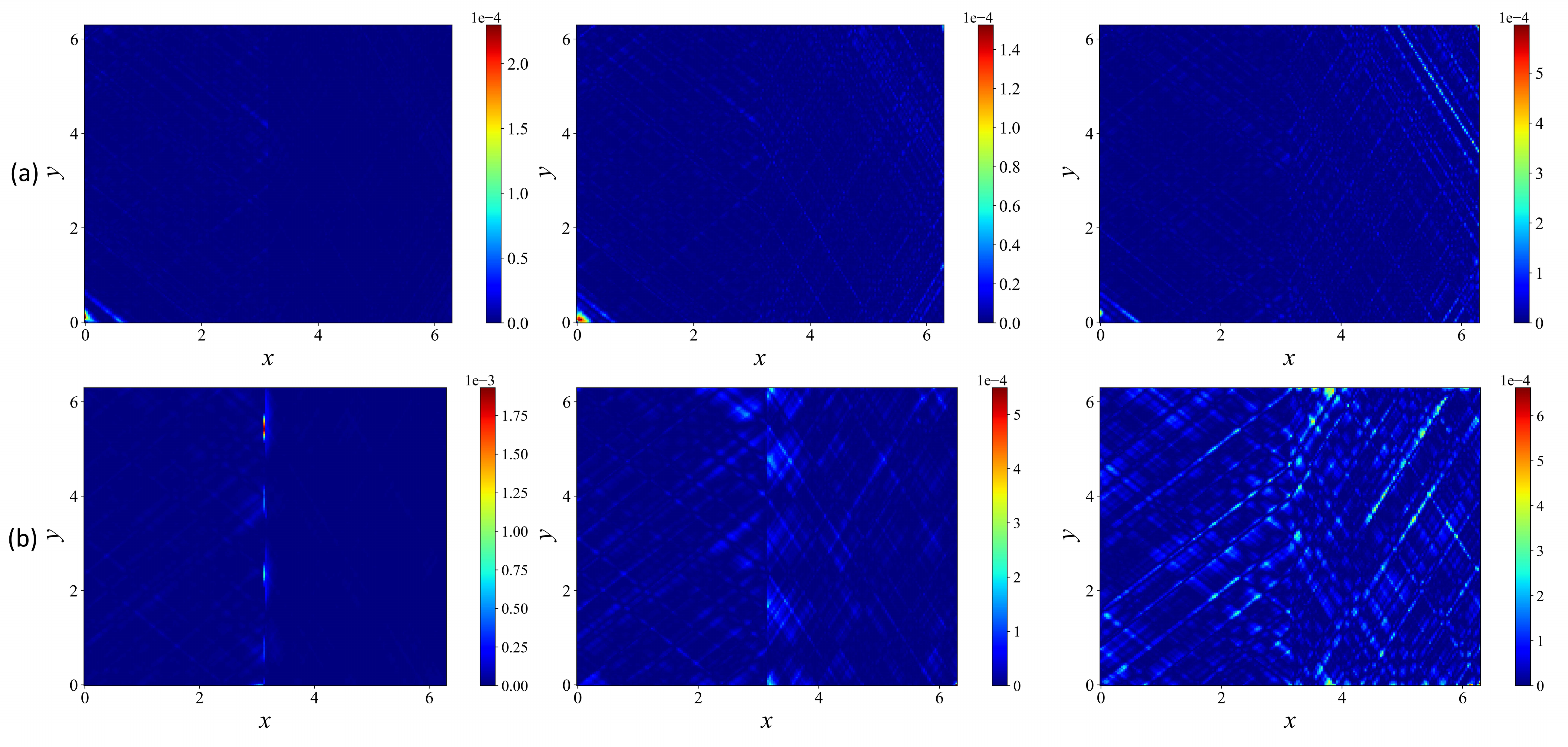}    
    \caption{Absolute errors for predictions. (a) $E_X$, $E_Y$, and $H_Z$ obtained by da-PINN.  (b) $E_X$, $E_Y$, and $H_Z$ obtained by PINNs.}
    \label{fig:2D}
    \end{center}
\end{figure}

\vspace{-1mm}
The two-dimensional Maxwell's equations under study are as follows:
\begin{equation}\label{eq:2dmaxwell}
    \begin{array}{ll}
        \hspace{-5mm}\displaystyle
        \varepsilon \dfrac{\partial E_{X}}{\partial t} = \dfrac{\partial H_{Z}}{\partial y},\hspace{2mm}
         \dfrac{\partial E_{Y}}{\partial x}\hspace{-0.8mm}-\hspace{-0.8mm}\dfrac{\partial E_{X}}{\partial y} \hspace{-0.8mm}=\hspace{-0.8mm} -\mu \dfrac{\partial H_{Z}}{\partial t},\\ [4mm]
        \hspace{-5mm}\displaystyle
  \varepsilon \dfrac{\partial E_{Y}}{\partial t} \hspace{-0.8mm}=\hspace{-0.8mm} -\dfrac{\partial H_{Z}}{\partial x},\hspace{1mm} \Omega\hspace{-0.8mm}=x\hspace{-0.8mm}\times\hspace{-0.8mm} y\hspace{-0.8mm}= \hspace{-0.8mm}\left[ 0,2\pi \right]\hspace{-0.8mm}\times\hspace{-0.8mm}\left[ 0,2\pi \right] ,  t\in \left[ 0,2 \right].
        \end{array}  
\end{equation}
The analytical solutions to \eqref{eq:2dmaxwell} are obtained according to \cite{NIU2022107906}:
\begin{equation}\label{eq:2dexa}
    \begin{array}{ll}
        \displaystyle
        E_{X}=\dfrac{k_{Y}}{\varepsilon \sqrt{\mu }\omega  } \mathrm{cos} \left ( \omega  t \right )
\mathrm{cos} \left ( k_{X}  x \right ) \mathrm{sin} \left ( k_{Y}  y \right ),\\[4mm] 
        \displaystyle
        E_{Y}=-\dfrac{k_{X}}{\varepsilon \sqrt{\mu }\omega  } \mathrm{cos} \left ( \omega  t \right )
\mathrm{sin} \left ( k_{X}  x \right ) \mathrm{cos} \left ( k_{Y}  y \right ),\\[4mm] 
        \displaystyle
        H_{Z}=\dfrac{1}{\sqrt{\mu }} \mathrm{sin} \left ( \omega  t \right )
\mathrm{cos} \left ( k_{X}  x \right ) \mathrm{cos} \left ( k_{Y}  y \right ),
        \end{array}  
\end{equation}
where $\omega= \dfrac{k_{X}^{2}+k_{Y}^{2}}{\mu\varepsilon}$. $\Omega _{1}=\left[0,\pi\right]\times \left[ 0,2\pi  \right]$ and 
 $\Omega _{2}=\left[\pi,2\pi\right]\times \left[ 0,2\pi  \right]$ are obtained with $d=\pi$ as the interface location.
The parameters in \eqref{eq:2dexa} are as follows:
\begin{equation}\label{eq:2dx}\notag
    \hspace{-2mm}
    \varepsilon\left ( \boldsymbol{x} \right ) =\begin{cases}
        2,  & \text{ if } \boldsymbol{x} \in \Omega _{1} \\
        5,   & \text{ if } \boldsymbol{x} \in \Omega _{2}
    \end{cases},
    \hspace{2mm}
    k_{X} \left ( \boldsymbol{x} \right ) =\begin{cases}
        2,  & \text{ if } \boldsymbol{x} \in \Omega _{1} \\
        4,   & \text{ if } \boldsymbol{x} \in \Omega _{2}
    \end{cases},
\end{equation}
$\mu=1$, $k_{Y}=2$, and $\boldsymbol{x}=\left[x,y\right]$. 
In this example, we set $N_{D}=8000$, $N_{P1}=N_{P2}=10000$, and $N_{I}=5000$.
There are $8$ hidden layers and $50$ neurons per layer in $Net_{i}$. We have tried ReLU and tanh as activation functions, in which ReLU with a better result is selected in this letter.
$\boldsymbol{\lambda}^{(0)}$ are randomly set as  $\mu_{1}^{(0)}=2$, $\varepsilon_{1}^{(0)} =1$, $\mu_{2}^{(0)}=13$, $\varepsilon_{2}^{(0)}=0$, and $d^{(0)}=15$. 
Table \ref{tb:2Dinversmax} shows the estimation performance.
Fig. \ref{fig:2dinverse} shows the comparison of da-PINN and PINNs, in which the estimations of da-PINN are more accurate near the interface.
Fig. \ref{fig:2D} shows the absolute errors of $E_{X}$, $E_{Y}$, and $H_{Z}$ obtained by da-PINN and PINNs, which indicates the relatively large prediction error of $E_{X}$ obtained by PINNs primarily happens near  the interface.
The $l_{2}\mbox{ relative errors}$ of predictions are listed in Table \ref{table:2dprel2}.
The better values are in the bold font  shown in Table \ref{table:2dprel2}.
\begin{table}[htb]
    \tabcolsep=0.5cm
	\renewcommand\arraystretch{1}
    \begin{center}
    \caption{Estimation performance of da-PINN for two-dimensional Maxwell's equations.}
    \vspace{1.0em}
    \setlength{\tabcolsep}{1.5mm}{
    \begin{tabular}{cccccc}
    \hline
    {\tabincell{c}{Parameters}} &
    \tabincell{c}{Estimations} &
    \tabincell{c}{$l_2$ relative error (\%)} \\\hline
    {\tabincell{c}{${\rm \mu_{1}}$}}  & 0.9997 & 0.0287  \\ 
    {\tabincell{c}{${\rm \varepsilon_{1}}$}}  & 2.0010 & 0.0501 \\ 
    {\tabincell{c}{${\rm \mu}_{2}$}}  & 1.0002& 0.0236 \\ 
    {\tabincell{c}{${\rm \varepsilon_{2}}$}}  & 4.9987& 0.0270 \\
    {\tabincell{c}{${d}$}}  & 3.1425 & 0.0274 \\\hline
    \end{tabular}}
    \label{tb:2Dinversmax}
    \end{center}
\end{table}
\begin{table}[H]
    \tabcolsep=1.5cm
	\renewcommand\arraystretch{1}
    \begin{center}
    \caption{Prediction performance of da-PINN and PINNs for two-dimensional Maxwell's equations.}
    \setlength{\tabcolsep}{1.5mm}{
    \begin{tabular}{c c c}
    \hline
    \tabincell{c}{Methods} &
    \tabincell{c}{Predictions} &
    \tabincell{c}{$l_{2}$ relative error}\\\hline
    \multirow{3}*{da-PINN}& {\tabincell{c}{$\widehat{E}_{X}$}} &  $\textbf{7.326}\boldsymbol{e}$-$\textbf{3}$  \\
    ~&  {\tabincell{c}{$\widehat{E}_{Y}$}} &  $\textbf{8.189}\boldsymbol{e}$-$\textbf{3}$  \\
    ~&  {\tabincell{c}{$\widehat{H}_{Z}$}} &  $\textbf{9.990}\boldsymbol{e}$-$\textbf{3}$   \\\hline
    \multirow{3}*{PINNs}&  {\tabincell{c}{$\widehat{E}_{X}$}} &  $1.870e$-$2$  \\
    ~&  {\tabincell{c}{$\widehat{E}_{Y}$}} &  $1.239e$-$2$  \\
    ~&  {\tabincell{c}{$\widehat{H}_{Z}$}} &  $9.605e$-$3$   \\\hline
    \end{tabular}}
    \label{table:2dprel2}
    \end{center}
\end{table}

\section{Conclusions}\label{se4}
In this letter, we proposed a da-PINN for solving the inverse problems of the Maxwell's equations in heterogeneous media and unknown interface location.
In da-PINN, two sub-networks are used to approximate the electric and magnetic fields of sub-domains. 
The parameters of the Maxwell's equations and interface location are incorporated into the loss function.
We also propose a domain-adaptive strategy to train da-PINN.
Experimental results verify the effectiveness of our method.

In the future, we will extend da-PINN to solve the Maxwell's equations in heterogeneous media with unknown shapes of interface.
This will enable da-PINN to solve more complex problems of the Maxwell's equations.

\bibliography{maxwell_inverse.bib} 
\end{document}